\pdfoutput=1

\documentclass[11pt]{article}
\usepackage{float}
\usepackage{amsmath}
\usepackage{amsfonts}
\usepackage[final]{acl}
\usepackage{times}
\usepackage{latexsym}
\usepackage{tikz}
\usetikzlibrary{positioning, arrows.meta, calc}
\usepackage[T1]{fontenc}
\usepackage{booktabs}
\usepackage{multirow}

\usepackage[utf8]{inputenc}

\usepackage{microtype}

\usepackage{inconsolata}

\usepackage{graphicx}

\title{LatentPrompt: Optimizing Promts in Latent Space}

\author{
  \textbf{Mateusz Bystroński\textsuperscript{1}} \And
  \textbf{Grzegorz Piotrowski\textsuperscript{1}} \And
  \textbf{Nitesh V. Chawla\textsuperscript{2}} \And
  \textbf{Tomasz Kajdanowicz\textsuperscript{1}}\AND
\\
\textsuperscript{1}Wrocław University of Science and Technology,
\\
\textsuperscript{2}University of Notre Dame
\\
\small{
\textbf{Correspondence:} \href{mailto:mateusz.bystronski@pwr.edu.pl}{mateusz.bystronski@pwr.edu.pl}
}
}

\begin{document}
\maketitle
\begin{abstract}
Recent advances have shown that optimizing prompts for Large Language Models (LLMs) can significantly improve task performance, yet many optimization techniques rely on heuristics or manual exploration. We present LatentPrompt, a model-agnostic framework for prompt optimization that leverages latent semantic space to automatically generate, evaluate, and refine candidate prompts without requiring hand-crafted rules. Beginning with a set of seed prompts, our method embeds them in a continuous latent space and systematically explores this space to identify prompts that maximize task-specific performance. In a proof-of-concept study on the Financial PhraseBank sentiment classification benchmark, LatentPrompt increased classification accuracy by approximately 3 percent points after a single optimization cycle. The framework is broadly applicable, requiring only black-box access to an LLM and an automatic evaluation metric, making it suitable for diverse domains and tasks.
\end{abstract}

\section{Introduction}

Large language models have demonstrated remarkable capabilities across tasks, but unlocking their full potential often hinges on the skillful design of prompts (natural language instructions). A well-crafted prompt can elicit more accurate and relevant responses from an LLM, whereas a suboptimal prompt may lead to poor performance. The art of prompt engineering has thus become critical for developing LLM-powered applications, yet it largely remains a manual ``trial and error'' endeavor. Practitioners must rely on intuition and heuristics to refine prompts, and even then, there is usually a lingering question: could the prompt be improved further?

The challenge of manual prompt engineering becomes particularly acute when deploying LLMs at scale or across diverse tasks. What works for one model or domain may fail for another, and the iterative process of refinement is both time-consuming and expertise-dependent. This has motivated researchers to explore automated techniques for prompt optimization that can systematically discover effective prompts with minimal human intervention.

Despite significant advances in automated prompt optimization, existing approaches face fundamental limitations. Methods requiring gradient access are restricted to white-box models, while black-box techniques often rely on discrete mutations that may miss subtle but effective variations. Current approaches also struggle to balance exploration of novel prompt formulations with exploitation of known effective patterns.

In our prior work, we introduced a latent-space ideation framework for generating creative ideas by navigating the continuous embedding space of concepts. That method demonstrated how textual outputs can be optimized in a semantic latent space and hinted at its generalizability to other NLP tasks. In this paper, we extend that framework to the realm of prompt optimization.

Our key insight is that prompts, being short texts, can be represented as vectors in a high-dimensional semantic space where different phrasing and structural variations correspond to continuous shifts. By moving through this latent space, we can discover prompt candidates that are semantically related to seed prompts but potentially more effective at eliciting desired behavior from LLMs. Our proposed latent space prompt optimization framework automatically generates variations of initial prompts and identifies those that yield improved task performance, requiring only a small validation set and treating the LLM as a black box.

To preview our results, we conducted a proof-of-concept experiment on financial sentiment classification using the Financial PhraseBank dataset \cite{malo2014financial}. Starting from seed prompts generated with GPT-4o to follow best prompt engineering practices, our framework autonomously generated candidate prompts by interpolating in the prompt embedding space, discovering a prompt that improved accuracy by nearly 3 percentage points over the best initial prompt.

\section{Related Work}

Automated prompt optimization has emerged as a critical research area, with approaches spanning multiple paradigms and optimization strategies. We organize existing work along several key dimensions: optimization space (discrete vs. continuous), optimization method, model requirements, and search strategy.

\subsection{Early Gradient-Based Approaches}

The first systematic approaches to automated prompt optimization leveraged gradient information to guide search in discrete token spaces. AutoPrompt \cite{shin2020autoprompt} pioneered this direction by using gradient-guided search to find effective discrete prompts for masked language models. The method computes gradients with respect to prompt tokens and selects replacements that maximize task performance. However, this approach requires white-box access to model gradients and was primarily demonstrated on smaller models like BERT, limiting its applicability to large-scale LLMs accessed through APIs.

Building on gradient-based foundations, recent work has explored more sophisticated optimization in continuous spaces. Wang et al. \cite{wang2025direct} presented methods that leverage gradient information in continuous embedding spaces while addressing instability caused by rounding continuous representations to discrete prompts. Their approach integrates greedy strategies with continuous optimization and exploits historical gradient information, showing improvements across models including GPT-2, OPT, Vicuna, and LLaMA-2.

\subsection{Reinforcement Learning Methods}

Reinforcement learning has provided another avenue for automated prompt optimization, treating prompt design as a sequential decision-making problem. RLPrompt \cite{deng2022rlprompt} optimizes discrete prompts using policy gradients, formulating prompt generation as selecting tokens from a vocabulary to maximize reward signals based on task performance. While effective at finding high-performing prompts, this approach often yields syntactically uninterpretable results that resemble ``gibberish tokens'' rather than human-readable instructions.

TEMPERA \cite{zhang2023tempera} addresses some limitations of prior RL approaches by focusing on test-time prompt editing with constrained action spaces. The method defines specific edit operations (insertion, deletion, substitution) and uses reinforcement learning to select sequences of edits that improve prompt performance. While this maintains better interpretability than unconstrained token generation, the predefined action spaces can limit exploration of truly novel prompt formulations.

\subsection{LLM-Based Optimization Frameworks}

A significant shift occurred with methods that leverage LLMs themselves as optimizers, treating prompt generation as a natural language task. Zhou et al. \cite{zhou2022ape} introduced the Automatic Prompt Engineer (APE), which uses large models to generate candidate instructions and selects among them based on validation performance. This approach maintains human readability while avoiding the need for gradient access, making it compatible with API-based models.

Promptbreeder \cite{fernando2023promptbreeder} extended this paradigm with a self-referential evolutionary strategy where an LLM mutates a population of prompts and evaluates their fitness on validation sets. The method gradually evolves better prompts over generations, demonstrating improvements over hand-crafted strategies like Chain-of-Thought prompting on reasoning benchmarks. However, the random mutation-based exploration can be inefficient without proper guidance mechanisms.

\subsection{Advanced Optimization Frameworks}

Recent work has introduced more sophisticated frameworks that combine multiple optimization strategies. Yang et al. \cite{yang2024opro} developed OPRO (Optimization by PROmpting), which treats LLMs as optimizers that iteratively improve prompts based on performance feedback described in natural language. This meta-optimization approach has shown effectiveness across various tasks but can be less effective with smaller-scale LLMs.

PromptWizard \cite{agarwal2025promptwizard} represents a significant advancement in feedback-driven optimization, employing a multi-agent framework with specialized roles for response generation, evaluation, and prompt refinement. The system incorporates expert-informed prompt principles as priors and maintains historical information about prompts, feedback, and scores. This structured approach to optimization has demonstrated substantial improvements, particularly for smaller and medium-sized language models.

\subsection{Continuous vs. Discrete Optimization}

The field has largely been divided between discrete optimization methods that work directly with natural language tokens and continuous methods that operate in embedding spaces. Discrete approaches like APE and Promptbreeder maintain interpretability but may miss subtle variations that could be effective. Continuous methods can explore smoother optimization landscapes but face challenges in mapping continuous representations back to coherent natural language.

Recent surveys \cite{cui2025survey, li2025survey} have identified this as a key frontier, with hybrid approaches showing promise for combining the advantages of both paradigms. However, most existing continuous methods still require gradient access or produce instability during the discrete conversion process.

\subsection{Limitations and Research Gaps}

Despite significant progress, current automated prompt optimization approaches face several fundamental limitations. Gradient-based methods are restricted to white-box models and often don't scale to the largest, most capable LLMs. Discrete search methods may miss effective prompt variations that require subtle semantic shifts rather than explicit token changes. Population-based approaches like evolutionary algorithms can be inefficient in high-dimensional prompt spaces without proper guidance.

Most critically, existing methods lack a principled way to navigate the semantic space of prompts. While discrete mutations can explore local variations and continuous methods can leverage smooth optimization landscapes, neither approach effectively captures the semantic relationships between different prompt formulations.

This gaps motivates our latent space approach, which aims to bridge continuous optimization with semantic meaningfulness in prompt design.

\section{Method}

\begin{figure*}[t]
    \centering
    \includegraphics[width=0.9\linewidth]{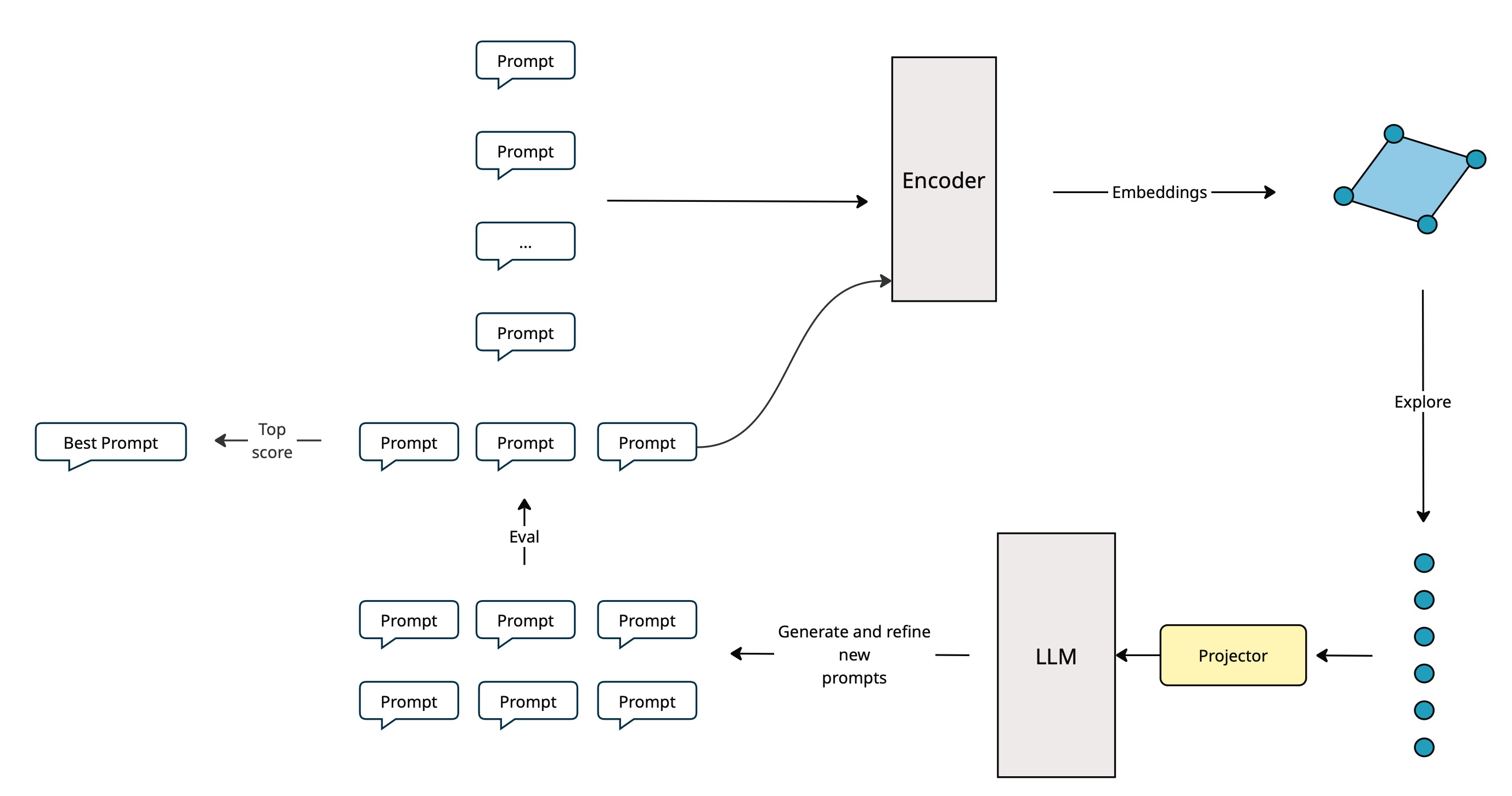}
    \caption{Overview of our framework. Seed prompt from user, or gathered using other techniques, are then encoded into embeddings via an encoder. The latent space formed by these embeddings is explored to produce new candidate vectors. These are projected into the token embedding space of a decoder LLM using an xRAG-style projector, enabling generation of new prompts. The generated outputs are refined and evaluated by an objective function and can optionally feed back into the prompt or embedding space, supporting iterative refinement.}
    \label{fig:architecture}
\end{figure*}

Our latent space prompt optimization framework adapts the architecture of latent-space ideation \cite{bystroński2025largelanguagemodelsinnovators} to the specific goal of improving instruction prompts. Figure 1 provides an overview of the system. Starting from one or more seed prompts (either supplied by a user or generated via an initial heuristic), the framework produces a set of new candidate prompts by exploring the semantic embedding subspace spanned by these seeds. It then evaluates each candidate and selects the top-performing prompts. The process can be iterated or refined as needed. The framework is modular and model-agnostic: different encoding models, language models, or evaluation functions can be plugged in without changing the overall pipeline. The core components of the framework include:

\paragraph{Prompt Encoder:} A text encoder that converts each prompt (instruction text) into a fixed-dimensional latent embedding. We denote a seed prompt as $p_i$ and its embedding as $e_i = \text{Enc}(p_i) \in \mathbb{R}^d$. The encoder can be any sentence or text embedding model that maps similar prompts to nearby vectors in semantic space. In our implementation, we use a pre-trained LLM-based encoder, treating it as a black-box feature extractor (no fine-tuning).

\paragraph{Latent Space Explorer:} A mechanism for generating new candidate embeddings from the initial prompt embeddings. Multiple strategies can be employed here:

\begin{itemize}
    \item \textbf{Interpolation:} Given two prompt embeddings $e_i$ and $e_j$, we generate a new point $e_{\text{new}} = \lambda e_i + (1 - \lambda) e_j$ for some $\lambda \in [0, 1]$. Interpolation creates a prompt that semantically blends the characteristics of two prompts. This was our primary strategy in experiments, using random pairs of seeds and $\lambda$ values near 0.5 for smooth blending.

    \item \textbf{Extrapolation:} We can set $\lambda$ outside the $[0,1]$ range (e.g., $\lambda > 1$ or $\lambda < 0$) to push beyond the convex combination of known prompts, potentially yielding more radical variations.

    \item \textbf{Noise Perturbation:} We can sample a random Gaussian perturbation vector $\epsilon \sim \mathcal{N}(0, \sigma^2 I)$ and add it to a seed embedding: $e_{\text{new}} = e_i + \epsilon$. This yields a prompt variant that is a random drift from a known prompt.
\end{itemize}

These operations allow exploring the neighborhood of known good prompts in a continuous way. In our current prototype, we focused on simple random interpolations between seed prompt embeddings, which already proved effective. More sophisticated exploration algorithms (e.g., Bayesian optimization or evolutionary search in embedding space) can be incorporated in this component to guide the search towards high-performing regions.

\paragraph{Cross-Modal Projection:} Since the candidate $e_{\text{new}}$ lives in the encoder’s embedding space, we need to translate it into a form that a language model can decode. We employ a learned linear projector $W_p: \mathbb{R}^d \rightarrow \mathbb{R}^m$, where $m$ is the dimension of the decoder LLM’s token embedding space. This yields $h = W_p(e_{\text{new}})$, a vector of the same dimension as the model’s word embeddings.

\paragraph{Prompt Decoder:} We then introduce $h$ to the LLM as a special pseudo-token. Essentially, we prompt the decoder to parahprase special token. This technique is following retrieval-augmented generation compression method \cite{cheng2024xrag}. It acts as a cross-modal bridge from the continuous latent space to the discrete language space.. One practical refinement we found helpful is to ensure the decoded prompt follows the stylistic and formatting conventions of effective prompts. Sometimes the raw decoded text might drift in format (e.g. missing an placeholder to insert input text). To address this, we included a secondary decoding step where we instruct the LLM to adjust the candidate prompt to match the format of the seed prompts, without changing its essential content.

\paragraph{Evaluator:} We evaluate each generated prompt by its actual performance on the target task. This requires a way to score a prompt $p$. In many settings, we can define an automatic objective metric. For instance, in classification tasks with a validation set, we can use the prompt to query the LLM on each validation example and measure accuracy against ground truth labels. More generally, the evaluator could be any function $f(p)$ that returns a numerical score indicating how well the prompt $p$ achieves the desired outcomes (e.g., BLEU or ROUGE for a summarization prompt, human preference scores for an open-ended generation prompt, etc.). In our implementation, we focused on classification accuracy on a validation dataset as the fitness measure for prompts. The evaluator thus runs the LLM with prompt $p$ on all validation inputs and computes the accuracy of the LLM’s outputs. Because running the LLM multiple times can be costly, we limited the size of the validation set used for evaluation. In principle, one could also train a cheaper proxy model to approximate the reward, or use an LLM-based evaluator that compares outputs to references, but in this work we use direct evaluation on ground truth since the task is well-defined.

\paragraph{Iterative Optimization:} The above components can be executed in a loop for iterative prompt refinement. After scoring the candidates, the top-performing prompts can be fed back in as new “seeds” to encode, and the process repeats to explore around this improved set. This allows a coarse-to-fine search where the prompt population hopefully converges toward an optimal region. In our experiments, we performed one main iteration (from initial seeds to first set of new prompts), but the framework is designed to support iterative evolution.

\section{Experiments}
We conducted a preliminary evaluation of our framework on a financial sentiment classification task using the Financial Phrase Bank dataset \cite{malo2014financial}. Our objective was to assess whether latent space exploration could discover prompts that elicit more accurate sentiment classifications from a large language model (LLM), compared to strong human-written baselines.

Our framework used \textit{Mistral 7B v2 }\cite{jiang2023mistral7b} as the prompt refirement and decoding model, \textit{SRF-Embeddings-Mistral} \cite{SFRAIResearch2024} as the embedding encoder, and the \textit{MLP Projector} architecture from~\cite{cheng2024xrag} to interpolate in latent space. For the downstream task, we also used \textit{Mistral 7B v2} as the task LLM responsible for predicting sentiment labels. To obtain the final label, we performed a second call to the LLM, as the output formats varied across prompts.

We began with five seed prompts, generated with GPT-4o prompted to follow best practices in prompt engineering. These prompts featured diverse styles such as role-based instruction, structured output formats, and step-by-step reasoning. Full prompt listings are provided in the Appendix.

We partitioned the dataset by extracting 10\% of the training split as a validation subset, used to score the generated prompts. The test split was reserved for the final evaluation. Prompt quality was assessed using accuracy, defined as the proportion of correct sentiment labels returned by the task LLM.

Using our framework, we performed a single latent space exploration step, generating 15 candidate prompts from the seed pool. The top 3 prompts were selected based on validation accuracy and subsequently evaluated on the test set.

Table~\ref{tab:prompt-results} compares the performance of the best seed prompt with the highest-scoring generated prompt:
\begin{table*}[h]
\centering
\begin{tabular}{ll}
\toprule
\textbf{Prompt} & \textbf{Accuracy (Test)} \\
\midrule
Best Seed Prompt (Baseline) & 75.36\% \\
Best Optimized Prompt (Ours) & \textbf{78.14\%} \\
\bottomrule
\end{tabular}
\caption{Accuracy of the top-performing seed prompt vs. the best prompt discovered via latent space optimization, on the financial sentiment classification task.}
\label{tab:prompt-results}
\end{table*}
These results demonstrate that latent space exploration can discover prompts that outperform strong hand-crafted instructions, even in a high-performance setting with modern instruction-tuned LLMs. This supports the hypothesis that prompt-space interpolation guided by task feedback is a viable strategy for prompt optimization.

\section{Discussion and Conclusion}

In this work, we presented a novel approach to automated prompt engineering by harnessing latent space exploration. Our latent space prompt optimization framework operates outside the LLM, treating it as an immutable component, and searches for an optimal way to query that model.

We avoided manual rewriting rules. Instead, the framework discovers prompt variants through vector arithmetic, making it adaptable to a wide range of tasks.

Because we do not require gradient access or fine-tuning, our method can be used with any proprietary or closed-source LLM (such as GPT-4 or other API-only models). We only need the ability to get model outputs for evaluation. This black-box nature makes the approach widely applicable.

The prompts generated by our method are valid natural language instructions, not cryptic token strings. This means humans can read, understand, and manually adjust them if needed. In high-stakes applications, having a human-readable prompt is important for transparency and trust. Our approach contrasts with pure continuous prompt tuning.

Our framework could be combined with discrete search or evolutionary algorithms. For example, one could use Promptbreeder's self-referential mutations~\cite{fernando2023promptbreeder} as part of generating seed variations, then refine those via latent interpolation. Or vice versa, use latent-generated prompts as initial populations for genetic algorithms. The continuous and discrete strategies are not mutually exclusive and could enrich each other~\cite{ju2023continuous, guo2023evoprompt}.

Our findings from the experiment indicate that latent space exploration is a viable tool for prompt optimization, but also highlight some challenges. One challenge is efficiency: evaluating a large number of candidate prompts can be costly since it involves many LLM queries. In our case, we mitigated this by using a small validation set and limiting candidates, but smarter search strategies could make the process more sample-efficient~\cite{prasad2024efficient}. Another challenge is maintaining prompt validity: as we saw, not all interpolated prompts were coherent.

The improvement we achieved via single iteration, 2.8 percentage points, while meaningful, was based on a relatively naive exploration. We suspect that with more advanced techniques, larger gains are possible.

In conclusion, latent space prompt optimization opens up a new avenue in the toolkit for aligning LLM behavior with user needs. It transforms the art of prompt crafting into a vector search problem that can be addressed with computational methods, without losing the natural language form of prompts. As LLMs become more central in AI deployments, the ability to auto-tune their prompts will be increasingly valuable for rapid development and adaptation to new tasks or domains.

\section{Future works}

As this is preliminary research, we are going to follow several directions.

Our current use of random interpolation barely scratches the surface of possible exploration methods. Future work will incorporate evolutionary strategies, swarm optimization techniques, or Bayesian optimization to more efficiently find high-performing prompts. These methods could guide the search by modeling the relationship between prompt embedding features and performance scores, rather than sampling blindly.

 Sometimes, we want prompts that not only maximize task accuracy, but also satisfy other criteria (minimal token length, certain style or politeness, avoiding particular phrases, etc.). The latent approach could be adapted to handle multi-objective optimization, by defining a combined score or using Pareto optimality to trade off different goals. For example, one could search for the shortest prompt that still achieves near-maximum accuracy.

 While we demonstrated the concept on a classification task, the framework should apply to a wide range of tasks. Each task may have different evaluation measures (e.g., F1 score, ROUGE, human preference ratings), but the loop of generate-and-evaluate remains the same. Conducting studies on diverse tasks will help validate the universality of the approach and might reveal task-specific tricks (for instance, prompts for reasoning vs. prompts for style transfer might occupy different subspaces).

Analyzing why certain directions in latent space correspond to effective prompt changes could deepen our understanding of prompt-language model interplay. For example, we could attempt to interpret the principal components or clusters of successful prompt embeddings to see what linguistic features they correspond to. This might lead to better initialization of the search (starting in regions known to be good, such as prompts that encourage step-by-step thinking for complex tasks).

Finally, our current evaluation is only a proof-of-concept. A key direction for future work is to perform a comprehensive evaluation, with thorough benchmarking across different tasks, datasets, and conditions to robustly assess the strengths and limitations of our approach.
\bibliography{custom}

\appendix
\section{Prompt Templates}
\label{appendix:prompts}

This appendix lists seed prompts used in our experiments and optimized prompt.

\subsection{Seed Prompts (GPT-4o Generated)}

\paragraph{Prompt 1}
\begin{quote}
\texttt{You are an expert linguist and sentiment analyst.} \\
\texttt{Task: Analyze the sentiment of the text provided below.} \\[0.5em]
\texttt{Text:} \\
\texttt{"\{text\}"} \\[0.5em]
\texttt{Instructions:} \\
\texttt{- Consider emotional tone, word choice, and implied intent.} \\
\texttt{- Categorize the sentiment as Positive, Negative, or Neutral.} \\
\texttt{- Justify your choice with a short explanation.} \\[0.5em]
\texttt{Output format:} \\
\texttt{\{"sentiment": "\textless Positive|Negative|Neutral\textgreater", "explanation": "\textless your reasoning\textgreater"\}}
\end{quote}

\paragraph{Prompt 2 (best seed)}
\begin{quote}
\texttt{Perform sentiment classification on the given user message.} \\[0.5em]
\texttt{Message:} \\
\texttt{"\{text\}"} \\[0.5em]
\texttt{Your response should include:} \\
\texttt{1. Sentiment label: Positive, Negative, or Neutral.} \\
\texttt{2. A list of words or phrases that influenced your decision.} \\[0.5em]
\texttt{Example output:} \\
\texttt{\{"label": "Positive", "evidence": ["love this product", "excellent quality"]\}}
\end{quote}

\paragraph{Prompt 3}
\begin{quote}
\texttt{Given the following customer feedback, identify the sentiment conveyed.} \\[0.5em]
\texttt{Feedback:} \\
\texttt{"\{text\}"} \\[0.5em]
\texttt{Requirements:} \\
\texttt{- Only choose one of: Positive, Negative, Neutral} \\
\texttt{- Provide a rationale in one sentence explaining your decision} \\[0.5em]
\texttt{Respond in this format:} \\
\texttt{Sentiment: \textless Positive|Negative|Neutral\textgreater} \\
\texttt{Reason: \textless brief explanation\textgreater}
\end{quote}

\paragraph{Prompt 4}
\begin{quote}
\texttt{Analyze the emotional tone of this sentence and label it accordingly.} \\[0.5em]
\texttt{Sentence:} \\
\texttt{"\{text\}"} \\[0.5em]
\texttt{Think step-by-step:} \\
\texttt{Step 1: Identify any emotionally charged words or phrases.} \\
\texttt{Step 2: Determine the speaker's intention or feeling.} \\
\texttt{Step 3: Classify the sentiment as Positive, Negative, or Neutral.} \\[0.5em]
\texttt{Final Output:} \\
\texttt{\{"sentiment": "\textless label\textgreater", "steps": ["\textless step 1\textgreater", "\textless step 2\textgreater", "\textless step 3\textgreater"]\}}
\end{quote}

\paragraph{Prompt 5}
\begin{quote}
\texttt{You are a language model trained to understand human emotions in text.} \\[0.5em]
\texttt{Input:} \\
\texttt{"\{text\}"} \\[0.5em]
\texttt{Your task:} \\
\texttt{- Classify the overall sentiment.} \\
\texttt{- Highlight if the sentiment is explicit (clearly stated) or implicit (implied).} \\
\texttt{- Return the result in the JSON format below:} \\[0.5em]
\texttt{\{"sentiment": "\textless Positive|Negative|Neutral\textgreater", "type": "\textless explicit|implicit\textgreater", "evidence": "\textless short quote or phrase from the text\textgreater"\}}
\end{quote}

\subsection{Optimized Prompt}

\begin{quote}
\texttt{Analyze the emotional tone of the given text and label it accordingly.} \\[0.5em]
\texttt{Text:} \\
\texttt{\{text\}} \\[0.5em]
\texttt{Instructions:} \\
\texttt{Step 1: Identify any emotionally charged words or phrases.} \\
\texttt{Step 2: Determine the sentiment as Positive or Negative.} \\[0.5em]
\texttt{Output:} \\
\texttt{\{"sentiment": "label", "steps": ["step 1", "Sentiment Analysis"]\}}
\end{quote}

The optimized prompt is only slightly different from Prompt 4, yet this small change leads to a noticeable performance difference.

\end{document}